\documentclass[letterpaper, 10 pt, conference]{ieeeconf}  

\IEEEoverridecommandlockouts                              

\usepackage{graphicx, caption} 
\usepackage{amsmath} 
\usepackage{amssymb}  
\usepackage{xcolor}
\usepackage{float}
\usepackage{algorithm}
\usepackage{algorithmic}
\usepackage{caption, subcaption}
\usepackage{cite}

\title{\LARGE \bf
Integrating Field of View in Human-Aware Collaborative Planning
}

\author{Ya-Chuan Hsu$^{1}$, Michael Defranco$^{1}$, Rutvik Patel$^{1}$, and Stefanos Nikolaidis$^{1}$
\thanks{$^{1}$Ya-Chuan Hsu, Michael Defranco, Rutvik Patel, and Stefanos Nikolaidis is with the Thomas Lord Department of Computer Science, University of Southern California, Los Angeles, CA 90089, USA
        {\tt\small yachuanh@usc.edu, michael.defranco@me.com, rutvikra@usc.edu, nikolaid@usc.edu}.}%
}

\begin{document}

\maketitle
\thispagestyle{empty}
\pagestyle{empty}

\begin{abstract}

In human-robot collaboration (HRC), it is crucial for robot agents to consider humans' knowledge of their surroundings. In reality, humans possess a narrow field of view (FOV), limiting their perception. However, research on HRC often overlooks this aspect and presumes an omniscient human collaborator. Our study addresses the challenge of adapting to the evolving subtask intent of humans while accounting for their limited FOV. We integrate FOV within the human-aware probabilistic planning framework. To account for large state spaces due to considering FOV, we propose a hierarchical online planner that efficiently finds approximate solutions while enabling the robot to explore low-level action trajectories that enter the human FOV, influencing their intended subtask. Through user study with our adapted cooking domain, we demonstrate our FOV-aware planner reduces human's interruptions and redundant actions during collaboration by adapting to human perception limitations. We extend these findings to a virtual reality kitchen environment, where we observe similar collaborative behaviors.
\end{abstract}

\section{INTRODUCTION}
\label{sec:intro}
Our work addresses the critical knowledge gap caused by sensing limitations in real-time collaborative planning, particularly the human field of view (FOV). In everyday tasks, humans naturally adapt their actions based on what others can see. For example, a driver will position themselves within another driver’s FOV to signal intent during lane changes, or a kitchen worker might place a plate within a chef’s FOV to communicate its readiness without interrupting their task. In such fast-paced collaborations, agents must coordinate actions within limited time frames. The success of the collaboration often hinges on whether all participants have access to the same information, which can vary due to FOV or the environment itself. This motivates the need to actively account for the human's understanding of the environment when selecting robot actions~\cite{shvo2023proactive, shvo2022resolving}. 

Supported by study insights on human behavior~\cite{koster2013theory}, human FOV significantly shapes their knowledge base (KB)~\cite{graf2022toward}, which in turn influences intentions and leads to varying behaviors. In collaborative tasks, anticipating human intentions, rather than merely reacting to observed behaviors, enables more seamless coordination. Prior works that consider FOV do not explicitly integrate intention reasoning into planning~\cite{sisbot2007human} and instead focus on broader situational awareness~\cite{shekhar2024human}, without incorporating the human's approximated 120-degree binocular FOV~\cite{howard1995binocular}. Both aspects are crucial for proactive collaboration in fast-paced settings, and our method directly addresses these gaps. We focus on collaborative scenarios where human intentions shift dynamically based on their KB and the progress of other agents. Specifically, we examine cases where one human remains engaged in a task while another agent advances the collaboration, creating a KB gap—a discrepancy between the human’s perception and the actual state of the world. By incorporating FOV into planning, we more effectively mitigate KB gaps and enhance coordination.

Our work is motivated by the idea of reducing KB gap in human-robot collaboration (HRC) via robots entering the human FOV to provide crucial information. To explore the concept, we adapt the well-known HRC domain, Overcooked AI~\cite{carroll2019utility}, to create the Steakhouse domain, which focuses on collaborative steak cooking with stationary tasks for food preparation. In this domain, one human agent prepares ingredients at the chopping board, while the robot handles plate cleaning at the sink and monitors the steak. Such setup helps investigate instances where the KB gap increases as the human focuses on a stationary task for extended periods.

For efficient coordination in fast-paced scenarios, non-verbal cues are preferred~\cite{lee2023effect}. Specifically, robot maneuvers that allow the robot to enter the human FOV, update the KB, and influence the human’s decision on subsequent tasks while the robot progresses in its own task. 
By using the Partially Observable Markov Decision Process (POMDP) framework, the robot balances actions that advance the task with those that reduce human's KB gap. For instance, when the robot passes by the human with a clean plate, it updates the KB, prompting the human to proceed with the next task, such as preparing the steak after chopping.

Given KB, we maintain an additional copy of the world. Given the challenge of exponential state space growth in POMDPs, we introduce an hierarchical approach. This method allows our POMDP formulation to scale yet remain tractable while considering KB. 

To further evaluate our FOV-aware planner, we implemented the Steakhouse domain in virtual reality (VR), allowing experts to collaborate with the robot agent in an immersive environment. VR enables users to experience naturalistic FOV limitations while interacting with the robot. Users can move freely and use VR controllers to perform tasks such as picking up ingredients, chopping, and plating steak, closely replicating real-world kitchen dynamics, and experiencing KB gaps due to their FOV. Our VR implementation provides us insights into practical challenges and effectiveness of our approach in close to real-world scenarios.

Our key insight is that the KB gap, \textit{induced by FOV}, affects the effectiveness of HRC. Hence, our work contributes
\begin{itemize}
    \item a \textit{Steakhouse domain} designed to investigate KB gap caused by tasks that require standing in one place for extended periods;
    \item an \textit{FOV-aware planner} that addresses uncertainty regarding the human's intended subtask, with decisions informed by the KB;
    \item a \textit{hierarchical solver} that supports real-time, computationally efficient FOV-aware human-robot collaboration within the complex Steakhouse domain; and
    \item an \textit{open-source VR version of the Steakhouse domain~\footnote[2]{https://github.com/SophieHsu/3d-plan-eval}}, allowing research to explore KB gaps that naturally arise from the limitations of VR avatars.
\end{itemize}
\vspace{-0.5em}

\section{Problem Definition}
\label{sec:problem_setting}
Consider a human and an agent collaborating to complete a task, consisting of multiple subtasks, where the human FOV is limited. Both agents aim to complete the task efficiently, minimizing the total amount of time steps taken.

We model the human as a heuristic agent that myopically selects subtasks to complete the overall task. The human prioritizes collaborative progress by choosing a subtask based on its immediate availability, such as picking up meat when the grill is empty or gathering a missing ingredient. When multiple subtasks are available and offer equal value in advancing the task, the human selects one randomly. 

The KB is updated continuously as objects enter or leave FOV. When an object enters the FOV, KB is updated to reflect the latest status of that object, which remains in the KB with its most recent status while visible. When an object leaves the FOV, its last-seen status persists in KB. If the human later observes that the object is no longer present in its last-seen location, the object is removed from KB.
Additionally, to capture the delay between an object entering the FOV and the human acknowledging it, the KB updates only after the object has been within the FOV for a certain period of time. 

Note that human modeling is not a primary focus of our paper; rather, we use a predefined model to evaluate our planning approach. The model, which reflects realistic decision-making in fast-paced collaboration and is used in prior work~\cite{fontaine2021importance}, provides robust basis for method evaluation.

\begin{figure*}[ht!]
    \centering
    \includegraphics[width=.95\linewidth]{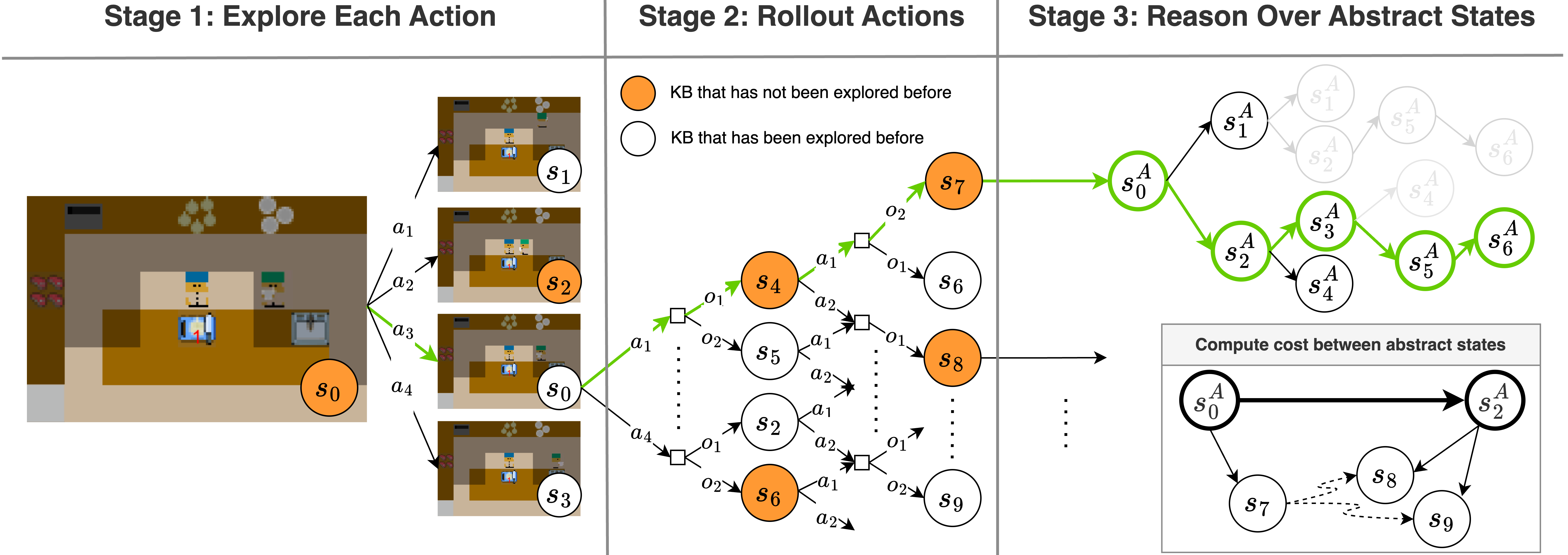}   
    \caption{Hierarchical online planning. We start by rolling out an action $a$ and obtaining new states (left). Each new state $s$ undergoes random exploration to obtain observations $o$ (center). Each rolled-out state (marked in orange), with a KB not seen in previously explored states, is reduced to an abstract state $s^A$. We then perform a look-ahead in the abstract state space (right). The costs between abstract states, shown in the lower graph, are computed by mapping abstract states back to the original state space. The planner ultimately chooses the action from the first stage that results in the highest value $V(s')$ (highlighted in green), indicating the optimal action.}
    \label{fig:method}
\end{figure*}

\section{Background}
\subsection{Knowledge-aware human-robot collaboration} 
Dating back to the early 2000s, literature on human-aware robot planning mentioned the concept of human knowledge~\cite{alami2006toward}. 
Over time, research has evolved: from fully cooperating based on human knowledge~\cite{zhang2017plan,dragan2013legibility,alami2006toward}, to planning optimally while taking human knowledge into account~\cite{chakraborti2017plan}, to revealing information to build human trust and cooperation~\cite{sreedharan2018hierarchical}, and finally, to influencing human perception during real-time collaboration~\cite{favier2023models, shekhar2024human, chakraborti2019explicability}.

We focus on real-time knowledge-aware collaboration planning, accounting for human FOV and KB gap. We assess 
collaboration behavior in our experiment setup, where KB gaps naturally arise due to limited FOV, between our FOV-aware planner and the human-controlled avatar.

\subsection{Partially Observable Markov Decision Processes}
We consider POMDPs, in which an agent takes actions to optimize reward by maintaining a belief over partially observable states. A POMDP is formally expressed as a tuple $(S, A, T, O, \Omega, R, \gamma)$, with states $S$, actions $A$, and observations $\Omega$. When the agent takes action \mbox{$a \in A$} in state \mbox{$s \in S$}, it moves to a new state \mbox{$s' \in S$} with probability $T(s, a, s') = p(s' | s, a)$ and receives an observation \mbox{$o \in \Omega$} with probability $O(s, a, o) = p(o|s, a)$. Since it does not know the true state, the agent uses its observations to construct a belief $b$, represented as a probability distribution over the states. As the agent action selection would be driven by goals, $R$ is a reward function of the state and, optionally, the action. $\gamma$ is a discount factor for future rewards.

Our work considers an online POMDP planner, which uses forward search in the belief space to locally approximate the optimal value function. 
In particular, we plan with QMDP~\cite{littman1995learning}, a popular method that estimates the expected values of actions based on the current belief and an assumption of no state uncertainty starting from the next time step. This assumption is sufficient for our approach, as we employ a few-step exploration rollout to capture potential new human behaviors resulting from entering FOV and changing KB (see Sec.~\ref{sec:outline}).
The $Q$ function is computed with the value function of the underlying MDP and optimized for a belief state $b$ as
\begin{equation}
\label{eq:pomdp_q_fn}
    Q(b,a) = \sum_{s\in S} b(s)\left(R(s) + \gamma \sum_{s'\in S}T(s,a,s')V(s')\right).
\end{equation}

\subsection{Hierarchical human-aware agent planning}
In long-horizon human-aware robot planning, where tasks consist of subtasks, solving large state spaces is a challenge. Hierarchical methods are effective for two reasons: first, they decompose long-horizon tasks into low-level motion planning and high-level decision-making for efficiency. Second, they align with human thinking, allowing abstract concepts like intent and preferences to be modeled by the symbolic planner, while the low-level planner manages motion control~\cite{singamaneni2023adaptive,silver2021learning,konidaris2018skills,lallement2018hatp}.

Hierarchy has been applied to POMDP~\cite{serrano2021knowledge}, while rewarding different levels in the hierarchy independently, resulting in not incorporating cross-layer information. 
A similar hierarchy approach is seen in \cite{ferrari2015hierarchical}, while they rely on explicit communication when intention diverges, which is less effective for fast-paced collaborations.

Our novel planner optimizes low-level actions and explores interactions within the human’s FOV, updating the KB. This approach integrates communicative values directly into low-level actions, allowing the robot to advance the task while making slight deviations to inform the human, avoiding complete interruptions with explicit communication. 
The abstract state search serves as a heuristic for efficient low-level action selection, integrating long-horizon planning effects.

\section{Approach via formulating as a POMDP}
\label{sec:model}
When the robot is uncertain about the human's intended subtask, specifically in cases that more than one feasible subtasks exist, stochasticity arises in human actions from the robot's perspective. To address this, we formulate the problem as a POMDP, enabling the robot to reason about the unobservable human subtask. Furthermore, this approach allows the robot to select actions that update the KB and subsequently influence the human's intended subtask, optimizing collaboration under the uncertainty in human behavior.

While our approach is general, we define it based on our collaborative cooking application as an example.
At a high level, our POMDP model considers a state space $S = \{S^W, S^R, S^H\}$, where $S^W$ represents information about the world, including whether the grill is empty or not, the onion is on the chopping board as a whole, or half chopped, or fully chopped; 
$S^R$ represents information of the robot, including its position, orientation, and held object; 
$S^H$ represents the information of the human, including the human's position, orientation, held object, KB and subtask. 

We assume the robot has full knowledge of $s^W \in S^W$, $s^R \in S^R$, and the human's position, orientation, and held object. These observable variables (also variables in the state space) constitute the observation space, $\Omega$, in POMDP.
With a given observation, the robot maintains a belief distribution $b$ over the human subtask, an unobservable variable.

The robot's objective is to complete all tasks with the least amount of actions taken. Consequently, we penalize the robot's number of steps by defining a negative reward for each step taken. We describe how we model the human and robot state dynamics in the subsections below.

\subsection{Robot state dynamics}
\label{sec:agent_dynamic}
The robot actions, including up, down, left, right, stay, and interact with objects, compose the action space, $A$, in our POMDP. 
The state of the robot, $s^R$, changes when the robot takes an action. 
For example, the robot's position, orientation, and held objects change based on actions selected.

\subsection{Human state dynamics}
\label{sec:subtask_dynamics}
The dynamics of the human state are modeled using a Markov chain, where each state in the chain encapsulates variables that describe the human’s current state. Within this structure, the available human subtasks are determined by the KB, which influences the possible actions within the Markov chain. 
For instance, given a KB with a washed plate in the sink and a cooked steak on the grill but no garnish prepared, the available human subtasks will include picking up the washed plate and picking up an onion. 
Another example is a KB where the robot holds an onion, the chopping board is empty, and the meat remains uncooked on the counter. Now, the only available human subtask is to pick up the meat.

The KB dynamics adhere to the rules specified in Sec.~\ref{sec:problem_setting}. In our application, an object must be in the human’s FOV for 3 timesteps before being acknowledged. For example, the initial KB may not include the robot's position or held object, but it updates once the robot remains in the human’s FOV for the specified duration.

\subsection{Human-Robot interaction}
\label{sec:human-agent_interaction}
The interactions between the human and robot are embedded into the transition functions. When the robot is not in the human FOV, $s^R$ and $s^H$ transition to their next state based on their individual dynamics. However, once the robot is in the human FOV, $s^R$ and $s^H$ are now tightly coupled. The KB in $s^H$ now becomes influenced by $s^R$ and the robot's action. 
For instance, if the robot picks up an object in the environment within the human FOV, the KB in $s^H$ is updated to the robot holding the object. 

\section{Hierarchical Planning}
\label{sec:outline}
We aim to achieve real-time collaboration in long-horizon tasks while incorporating FOV and KB. With the POMDP formation of our problem definition, we introduce an online hierarchical planner designed for real-time planning. 
The method aims to select the action with the maximum $Q$ value (Eq.~(\ref{eq:pomdp_q_fn})) by uniformly sampling possible next states to explore human FOV, abstracting the discovered states to reduce the dimension of the searched state space, and performing a heuristic search on the abstract state space. Below, we detail these processes.

\subsection{Random exploration}
\label{sec:kb_rollout}
The motivation behind this exploration is to seek out actions that enter human FOV and change the KB, leading to new next states $s'$ with higher $V(s')$. The exploration rolls out actions and observations as shown in Stage 2 in Fig.~\ref{fig:method}.
For our online planner, $T(s,a,s')$ is the probability of reaching state $s'$ starting from state $s$ after following a trajectory $\tau = \{a, a_1, ..., a_{n-1}, a_n\}$, where $a$ is the initial action and the trajectory comprises $n$ actions. We compute $T(s,a,s')$ based on the estimated likelihood of reaching $s'$ with randomly sampled trajectories $\tau$ from $s$.
Such a design captures the influence the trajectory has on the KB (as mentioned in Sec.~\ref{sec:human-agent_interaction}).

\subsection{Mapping to abstract state}
\label{sec:map_abstract}
To estimate the value of influencing the human's decision on the next subtask in terms of task progression, we seek long-horizon task rewards by searching the abstract space.

For each newly discovered $s'$ in Stage 2, we construct abstract states $s^A = f(s)$, where $f:S \rightarrow S^A$ is an abstraction function and $S^A \subset S$. 
Such a step is designed to handle large-state space POMDPs using abstract states to compute long-horizon plans. 
We construct the abstract state under the assumption that once the current human subtask is known from state $s$, the human’s FOV limitations are removed, and the KB accurately aligns with the actual environment.
We additionally assume the robot and human complete subtasks via the shortest path; therefore, we exclude the position and orientation information, capturing their influence in the cost function instead.
The remaining human-related variable in the abstract state space is the human subtask.

\subsection{Heuristic search on abstract states}
\label{sec:abstract_rollout}
We estimate the value of each abstract state (see Stage 3 of Fig.~\ref{fig:method}), with a heuristic function, defined as the sum of the negative values of the costs $C(s^A_i, s^A_{i+1})$ (see Fig.~\ref{fig:method}) for transitioning between abstract states during the executed trajectory and the immediate reward of reaching each abstract state $R(s^A_{i+1})$.
An example of cost given $s^A_i = \{s^W_i, s^R_{\text{hold nothing}}, s^H_{\text{wash a plate}}\}$ and $s^A_{i+1} = \{s^W_{i+1}, s^R_{\text{hold onion}}, s^H_{\text{pick up plate}}\}$ is the smallest number of steps required for either the robot to reach the onion station or the human to reach the sink.
An example of the reward will be a reward of $10$ for reaching the state with washed plates and $100$ for reaching a state that delivered a cooked steak on a clean plate. 
In summary,
\begin{equation}
    V(s') = -C(s^A_1, s^A_2)+R(s^A_2)-...-C(s^A_{n-1}, s^A_{n})+R(s^A_n),
\label{eq:abstracted_v_fn}
\end{equation}
where $n$ is the look-ahead length in the abstract state space.

\subsection{Online policy}
\label{sec:online_policy}
To seek optimal policy for the $Q$ function, we implement $T(s,a,s')$ following Sec.~\ref{sec:kb_rollout} for Eq.~(\ref{eq:pomdp_q_fn}), and with $V(s')$ computed based on Eq.~(\ref{eq:abstracted_v_fn}). 
We repeat the action exploration and abstract state search for each step and select $\mathrm{argmax}_{a\in A}Q(b,a)$.

\section{Experiment setup}
\label{sec:exp_setup}
\begin{figure}[h]
    \centering
    \includegraphics[width=0.8\linewidth]{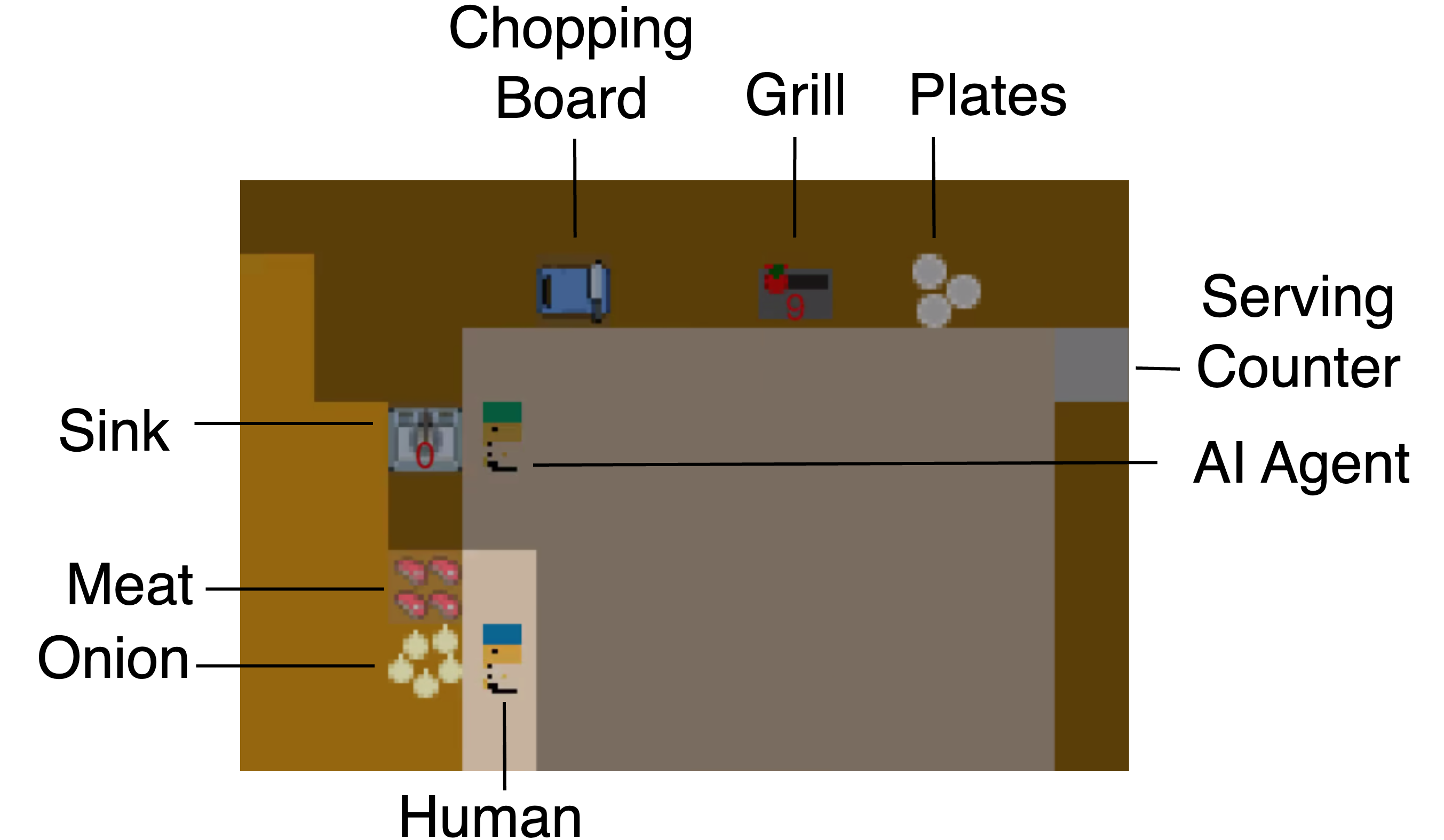}
    \caption{Steakhouse domain. (Dimmed tiles are completely black during user studies to simulate FOV.)}
    \label{fig:game_interface}
    \vspace{-1.5em}
\end{figure}

Our experiments aim to analyze the collaborative behavior when the robot considers the limited human FOV.
We first analyze simulated experiment results in the Steakhouse domain (shown in Fig.~\ref{fig:game_interface}).

\subsection{Steakhouse domain study}
As motivated in Sec.~\ref{sec:intro}, 
the two agents in the Steakhouse domain will cook the meat, chop the onions, and wash the plates for plating the cooked steak and chopped garnish. 
To represent the human FOV, the tiles outside of the human FOV are black during user studies.

\subsection{Human model}
\label{sec:human_model}
We use a human model from prior work~\cite{carroll2019utility}, which selects the subtask that immediately contributes to progressing the collaboration based on the current world state. The model does not optimize for the horizon of subtasks or the robot's actions. Following Sec.~\ref{sec:subtask_dynamics}, we obtain a set of available subtasks and select the next one based on the human model.
Once the subtask is decided, the human will take the shortest path to complete it, computed using $\mathrm{A}^*$.

\subsection{Robot planner}
\label{sec:exp_agent_planner}
We apply the formulation of our method as the illustrated example in Sec.~\ref{sec:model}. Abstract states capture essential information for collaboration tasks, including the status of the grill, sink, chopping board, and the number of orders remaining.

For forming the abstract state, we define the human’s internal variable as the subtask they aim to complete, which in our cooking domain includes tasks such as picking up meat, chopping onions, washing plates, plating, and serving steak. Since the subtask is unobservable, we maintain a belief distribution over the abstract state space throughout the collaboration. This belief is updated based on observable information, including human’s current position, heading direction, and KB. The belief in a specific subtask increases when the human’s heading direction aligns with the subtask’s destination, the distance to the subtask decreases, and is feasible based on the KB.

We implement two robot planners, each with a different human model. The FOV-aware robot uses a model with FOV limited to 120 degrees, while the baseline robot employs a model with full perceptual capability. Both planners considers rollouts of human trajectories during planning.

\subsection{Qualitative analysis}
\label{sec:2d_qualitative}
We analyze behaviors using the same simulated human model across different robot planners and kitchen layouts, designed to reflect common household configurations (e.g., island-centered, Peninsula kitchens). Each behavior highlights how the FOV-aware robot deliberately positions itself within the human FOV to influence KB.

\textbf{Behavior scenario 1: Robot walks alongside the human.}
\begin{figure}
    \centering
    \includegraphics[width=0.47\textwidth]{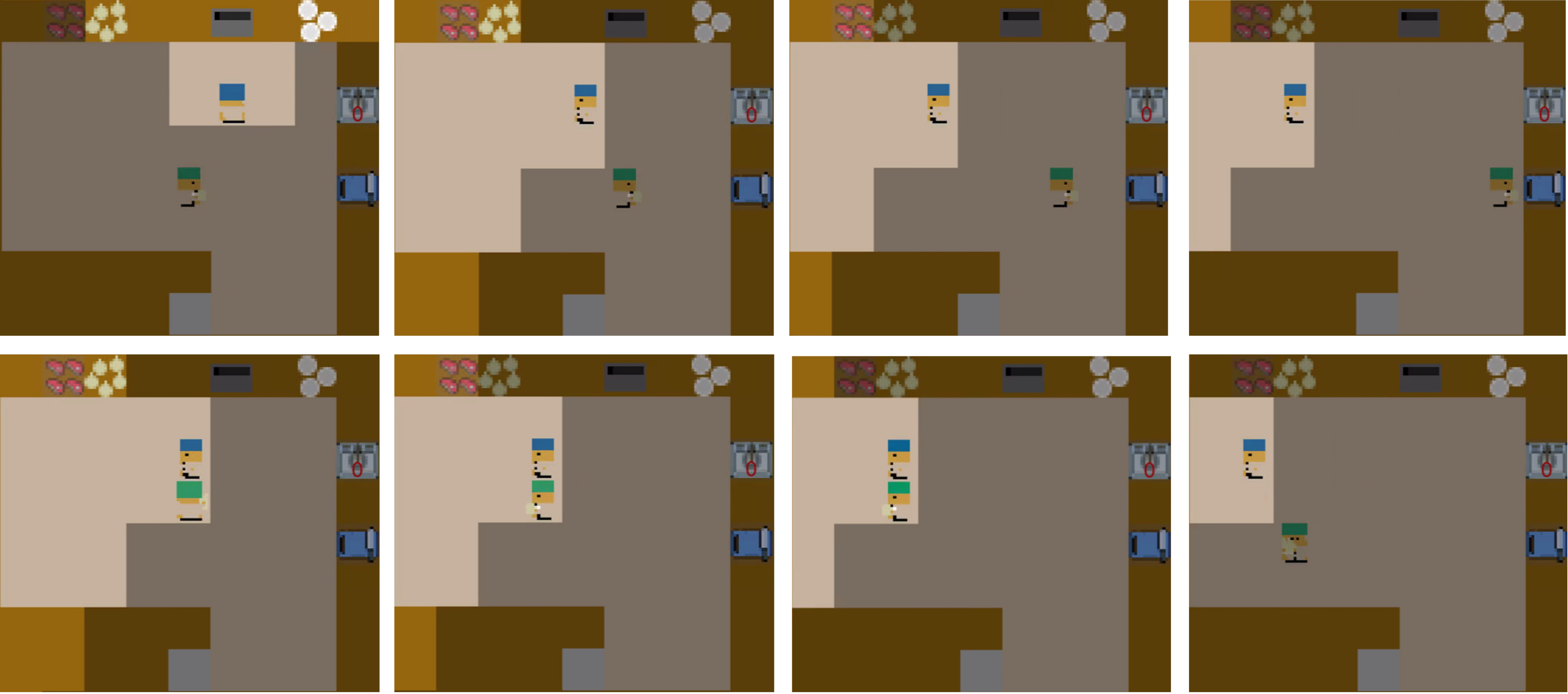}
    \caption{Collaboration behavior in a Peninsula kitchen. Top row: the baseline robot placing an onion directly on the chopping board; Bottom row: the FOV-aware robot revealing the onion to the human before proceeding. (Dimmed tiles are black during user studies to simulate FOV.)}
    \vspace{-0.5em}
    \label{fig:side2}
\end{figure}
In the top row of Fig.~\ref{fig:side2} (Peninsula kitchen), the baseline robot (green agent) places an onion on a chopping board without the human noticing. In the bottom row, the FOV-aware robot deliberately deviates from its initial path, walking alongside the human for 3 timesteps to ensure the human sees it holding the onion.

\textbf{Behavior scenario 2: Robot positions to stand in front of workstations.}
\begin{figure}
    \centering
    \includegraphics[width=0.47\textwidth]{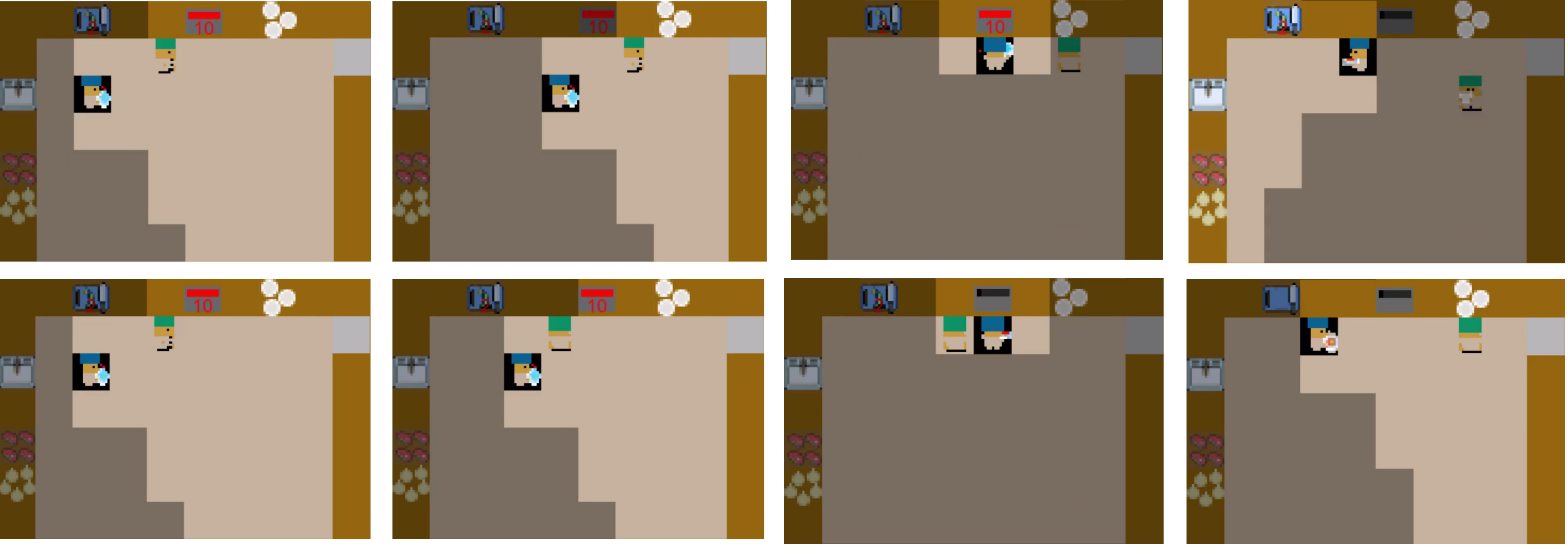}
    \caption{Collaboration behavior in a $\cap$-shaped kitchen. Top row: the baseline robot (green agent) picks up a plate assuming the human is aware. Bottom row: the FOV-aware robot waits for the human to pick up the cooked steak. (Dimmed tiles are black during the study to simulate FOV.)}
    \vspace{-1.5em}   
    \label{fig:none3}
\end{figure}
We illustrate how the FOV-aware robot delays its action to avoid performing tasks outside the human FOV. Both the FOV-aware and the baseline robot start at the same position in Fig~\ref{fig:none3} ($\cap$-shaped kitchen), and both intend to head to the plate station.
In the baseline robot scenario, the robot walks directly to the plate station and picks up a plate.
The FOV-aware robot, however, turns north and stays there while the human passes by and travels toward the grill.
It then proceeds to its goal location while the human turns around to deliver the completed dish.
This decision-making pattern prevents the human from not knowing the plate has already been picked up.

\begin{figure}[h]
    \vspace{-0.5em}   
    \centering
    \includegraphics[width=0.49\textwidth]{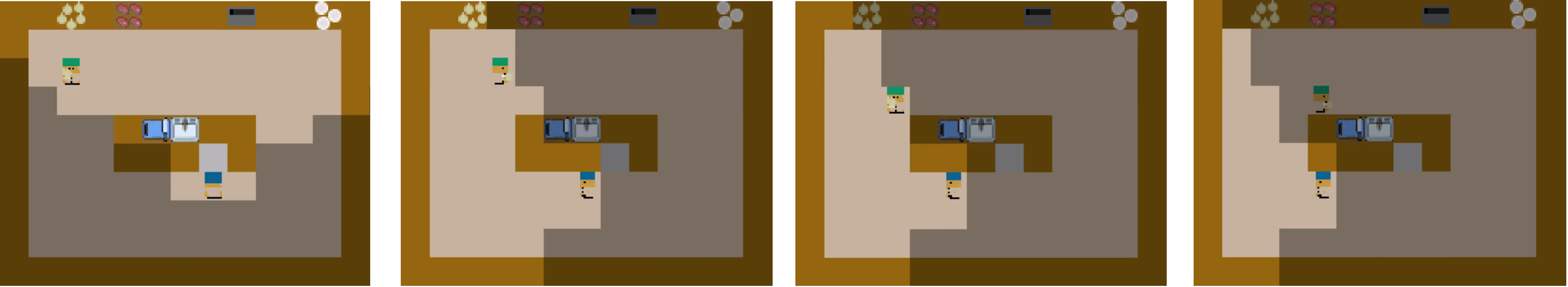}
    \caption{The FOV-aware robot (green agent) follows a staircase-like trajectory to remain within the human FOV.}
    \label{fig:mid2_ex2}
    \vspace{-0.5em}   
\end{figure}

\textbf{Behavior scenario 3: Robot aligns its path with the human FOV.}
We showcase the FOV-aware robot planning its path based on the human FOV. 
In Fig.~\ref{fig:mid2_ex2} (island-centered layout), the FOV-aware robot moves within human FOV with a staircase-wise trajectory to show it is holding the onion. 

\subsection{Quantitative analysis}
We simulate the two planners across $6$ different layouts, each tested $10$ times, and measure total low-level actions executed for task completion. We observe that the FOV-aware planner completed within an average of $152.1 \pm 9.4$ actions, as opposed to $165.1 \pm 10.9$ actions of the baseline planner.

\section{User Study}
We investigate whether collaboration differences observed in simulation persist when a simulated robot interacts with real users, bridging the gap between simulation and real-world interaction.
We conducted a user study ($n=27$), where participants collaborated with different robot planners in a $\cap$-shaped kitchen (Fig.\ref{fig:none3}) and an island-centered layout (Fig.\ref{fig:mid2_ex2}). To prevent bias, the order of robot planners and layouts was counterbalanced across participants.

\subsection{User interface design}
\label{sec:user_interface}
Participants select subtasks during collaboration, while human movement follows an A*-computed shortest path.
This design choice eliminates noise introduced by the diverse ways humans navigate to complete a subtask, allowing the study to focus on understanding how influencing the KB gap, assessed by comparing the KB with the world state, can change the decision-making of the human’s next subtask and improves collaboration. 
In summary, we assume the robot accurately understands the human motion model.

Our interface also supports interrupt actions, including movement (up, down, left, right), stay, and interact. These allow participants to halt subtasks, make one-step adjustments, or correct actions due to environmental misperceptions. For instance, if the participant picks up meat assuming that the grill is free but finds it occupied, they may use interrupt actions to place the meat on an empty counter. We capture these interrupt actions as they indicate decision-making shifts.

\subsection{Hypotheses}
\textbf{Hypothesis 1.} \textit{In the experiment, \textbf{the KB gap will be smaller} when the human collaborates with the FOV-aware robot.} 
KB gap is the differences between the KB and the current world state. For example, if the KB shows the onion is not chopped, but the world state shows it is chopped, the KB gap is 1. 
We expect the FOV-aware robot to reason over the human's limited FOV, enabling more accurate KB updates and more effective task performance. 

\textbf{Hypothesis 2.} \textit{The human performs \textbf{fewer interrupt actions} when collaborating with an FOV-aware robot.} 
The more informed human are of the actual environment, the less likely they perform subtasks that later become redundant and change their selected subtasks during execution. This leads to reducing the need for interrupt actions (defined in Sec.~\ref{sec:user_interface}) to undo the chosen subtask.

\subsection{Results}
We evaluate the impact of robot type and kitchen layout (independent variables) on KB gap and interruption frequency (dependent variables).
A two-way repeated ANOVA revealed a significant interaction between the independent variables on KB gap (F(1,27)=4.526, $p$=.043) and interruption frequency (F(1,27)=4.692, $p$=.039).

Given the significant interactions among independent variables, we analyzed simple main effects across layouts. 
In the $\cap$-shaped kitchen, FOV-aware robots showed a smaller KB gap than baseline robots (F(1,27)=6.705, p=.015, $\eta^2G$ =0.111). In contrast, in the island-centered layout, the data did not provide sufficient evidence of a statistically significant difference (F(1,27)=0.05, $p$=.823, $\eta^2G$=0.001). 
Regarding interruption frequency, working with an FOV-aware robot led to significantly fewer interruptions in the $\cap$-shaped kitchen (F(1,27)=7.192, $p$=.012, $\eta^2G$=0.088), while the differences found in the island-centered layout were not statistically significant (F(1,27)=0.228, $p$=.636, $\eta^2G$=0.003).  

We attribute the non-significant findings in the island-centered layout to the fact that the human can easily observe the environment. For example, when the human delivers a prepared dish to the serving counter located on the center kitchen island, all objects come within the human FOV.

\section{Virtual-Reality Demonstration}
In contrast to the experiment setup in Sec.~\ref{sec:exp_setup}, by evaluating collaboration behavior in an immersive VR environment, we can naturally replicate human FOV limitations and the time required for stationary tasks, such as chopping ingredients. Such setup leads KB gap created under realistic conditions, allowing for a more authentic assessment of human-robot collaboration. The VR environment effectively captures the dynamic nature of these limitations, providing valuable insights into how KB gaps influence decision-making and task efficiency during collaboration.

\subsection{Virtual reality simulation setup}
\begin{figure}[htbp]
  \begin{minipage}{0.23\textwidth}
    \centering
    \includegraphics[width=\textwidth]{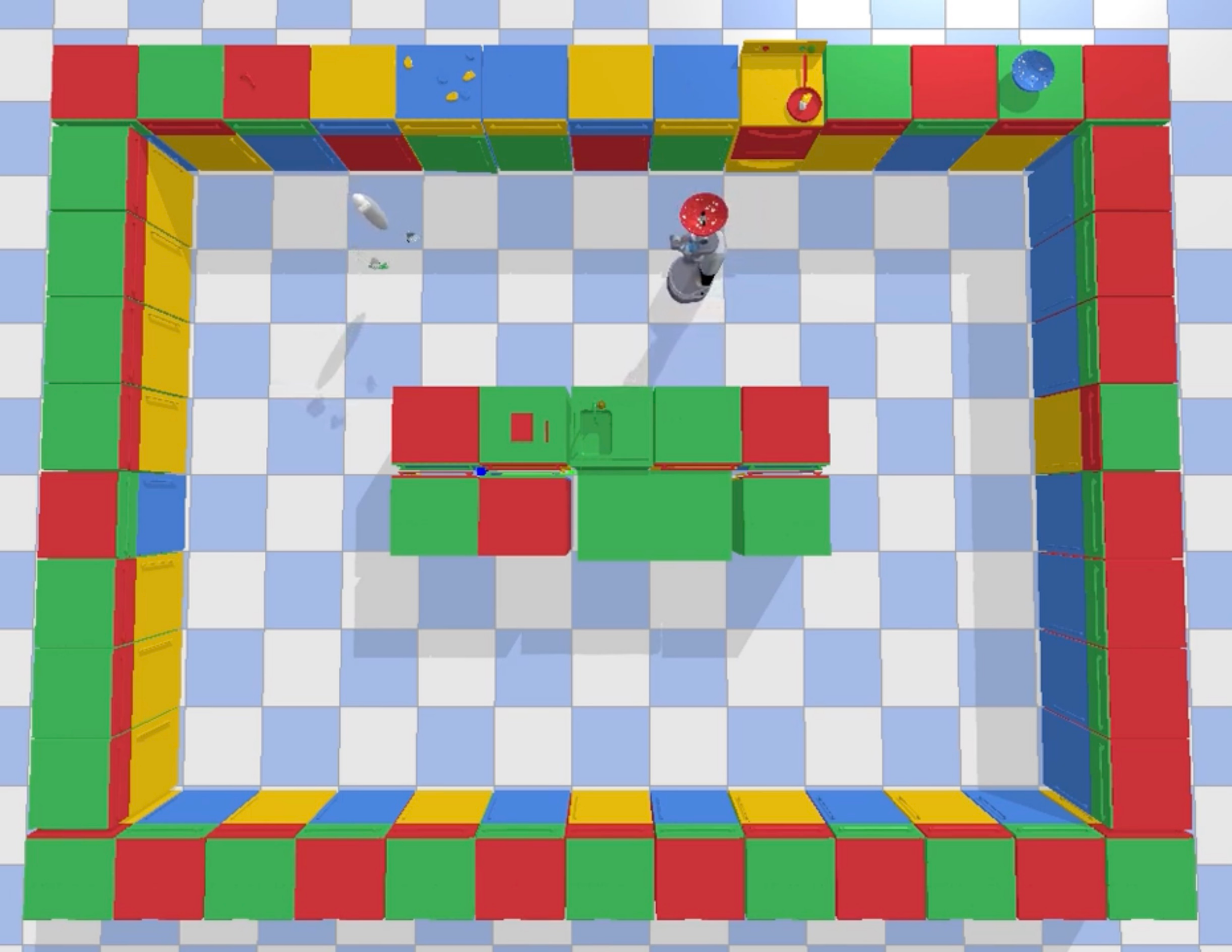}
    \caption{VR kitchen}
    \label{fig:vr_grid}
  \end{minipage}
  \begin{minipage}{0.23\textwidth}
    \centering
    \includegraphics[width=\textwidth]{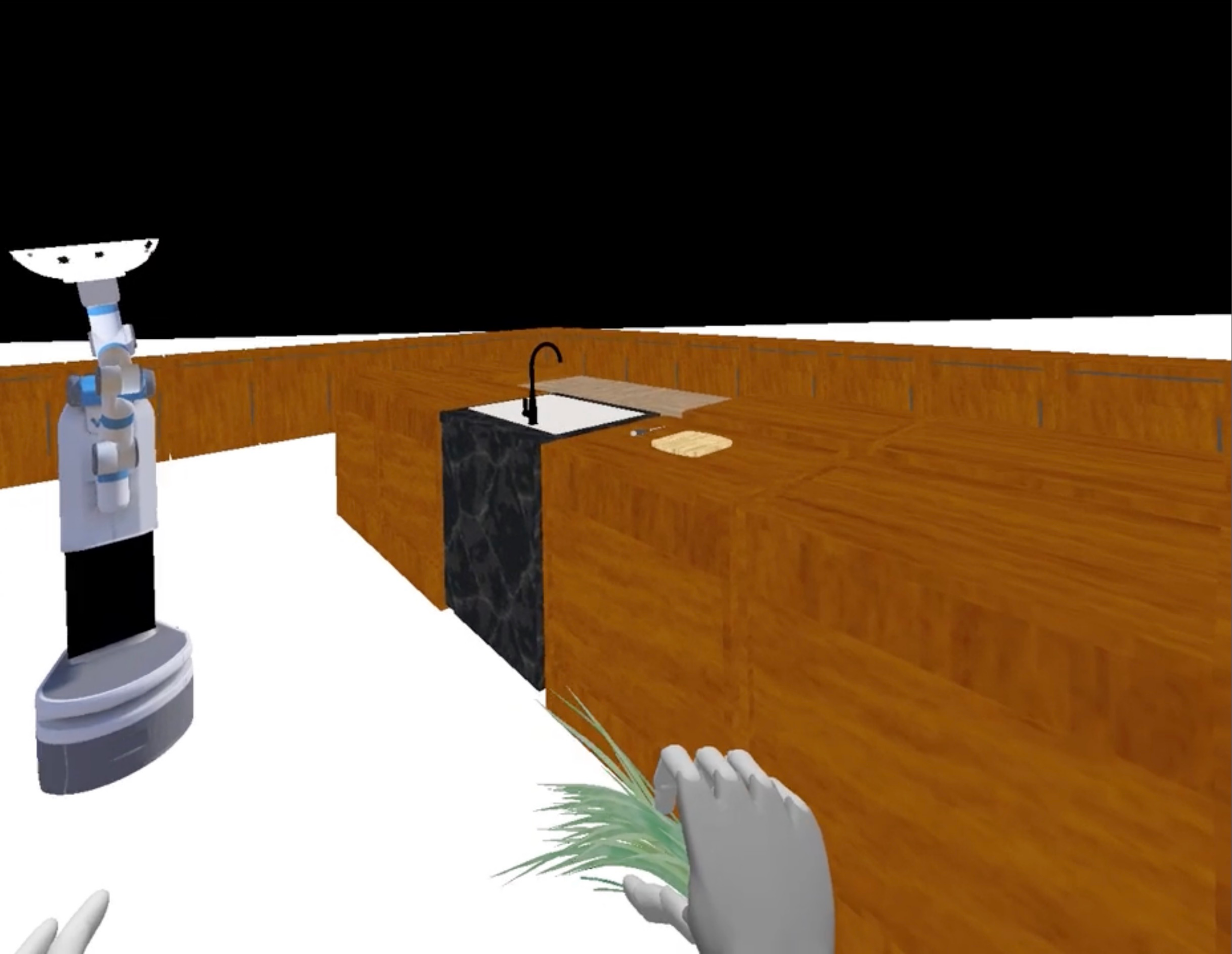}
    \caption{VR first-person view}
    \label{fig:vr_fpv}
  \end{minipage}
    \vspace{-1.5em}   
\end{figure}

The VR kitchen (Fig.\ref{fig:vr_grid}) was developed with iGibson~\cite{li2022igibson}, which supports dynamic object states (e.g., onions can be whole or chopped). We used a Meta Quest 2 headset with handheld controllers for object interaction, and navigation was handled via the left-hand toggle while users remained seated in a rotating chair for safety. 

To assess our algorithm with VR, human experts controlled the VR avatar and interacted with a robot in a rule-based game-play manner (Sec.~\ref{sec:human_model}).
Given VR is an immersive environment, the FOV limitation is a ``built-in feature'' (Fig.~\ref{fig:vr_fpv}); hence, the experts navigated the VR kitchen and selected the next subtask based on their KB.

\subsection{Qualitative analysis}

We observed two types of behavior patterns$^{2}$.

\textbf{Behavior 1: Robot prolongs its visibility to the human.} The FOV-aware robot deliberately takes additional steps to remain within human FOV. A strategy observed for extending visibility is the robot staying in place to remain in the human FOV for at least 3 timesteps, allowing the human to notice the item it picked up before leaving the human FOV.

\textbf{Behavior 2: Robot chose to take a longer path to enter the human FOV.}
In the island-centered layout, the FOV-aware robot takes a longer path around the kitchen island to deliver a finished dish, ensuring it enters and remains within the human FOV. In contrast, the baseline robot follows a shorter path to deliver the dish. Additionally, the FOV-aware robot waits for the human to complete their task and turn around before retrieving the garnish from the chopping board. The baseline robot, on the other hand, picks up the garnish without waiting for the human to turn around.

\section{Conclusion}
We introduced a real-time, FOV-aware planning approach that reduces redundant participant movements, with the main strength being to proactive adapt robot trajectories to keep humans informed without sacrificing overall task performance. 
However, due to approximations in real-time action computation, such as abstract state spaces and heuristic-based value estimation, participants sometimes perceived the robot’s actions as sub-optimal or confusing. By highlighting these strengths and limitations, we aim to inspire further research on environments where human perceptual constraints influence collaboration.

\section{Acknowledgment} 
This work was partially supported by NSF NRI \#2024936 and the Agilent Early Career Professor Award. We thank Jason Feng and Sujay Garlanka for contributing to the open-source VR version of the Steakhouse domain.

\bibliographystyle{IEEEtran}
\bibliography{IEEEabrv,main}

\end{document}